\documentclass[conference]{IEEEtran}
\IEEEoverridecommandlockouts

\usepackage{cite}
\usepackage{amsmath,amssymb,amsfonts}
\usepackage{graphicx}
\usepackage{textcomp}
\usepackage{xcolor}
\usepackage{booktabs}
\usepackage{array}
\usepackage{multirow}
\usepackage{url}
\usepackage[hidelinks]{hyperref}

\hypersetup{
    pdfauthor={Changyu Lee, Yeonsoo Park, Abdullah Alfarrarjeh, Seon Ho Kim},
    pdftitle={What Does CLIP Learn for Regional Geolocalization? Probing Visual Cues and Scene Configuration After Adaptation},
    pdfsubject={Fine-grained urban geolocalization with CLIP},
    pdfkeywords={visual geolocalization, CLIP adaptation, spatial configuration, visual cue dependence}
}

\begin{document}

\title{What Does CLIP Learn for Regional Geolocalization? Probing Visual Cues and Scene Configuration After Adaptation}

% \author{
% \IEEEauthorblockN{Changyu Lee}
% \IEEEauthorblockA{
% \textit{University of Southern}\\
% \textit{California}\\
% Los Angeles, USA\\
% clee1806@usc.edu
% }
% \and
% \IEEEauthorblockN{Yeonsoo Park}
% \IEEEauthorblockA{
% \textit{Integrated Media Systems}\\
% \textit{Center (IMSC)}\\
% \textit{University of Southern}\\
% \textit{California}\\
% San Francisco, USA\\
% tripwithsoo@gmail.com
% }
% \and
% \IEEEauthorblockN{Abdullah Alfarrarjeh}
% \IEEEauthorblockA{
% \textit{German Jordanian}\\
% \textit{University}\\
% Amman, Jordan\\
% abdullah.alfarrarjeh@\\
% gju.edu.jo
% }
% \and
% \IEEEauthorblockN{Seon Ho Kim\textsuperscript{*}}
% \IEEEauthorblockA{
% \textit{Integrated Media Systems}\\
% \textit{Center (IMSC)}\\
% \textit{University of Southern}\\
% \textit{California}\\
% Los Angeles, USA\\
% seonkim@usc.edu
% }
\author{
\IEEEauthorblockN{
Changyu Lee\textsuperscript{1},
Yeonsoo Park\textsuperscript{1},
Abdullah Alfarrarjeh\textsuperscript{2},
Seon Ho Kim\textsuperscript{1,*}
}

\IEEEauthorblockA{
\textsuperscript{1}\textit{University of Southern California, Los Angeles, USA}\\
\textsuperscript{2}\textit{German Jordanian University, Amman, Jordan}\\
\{clee1806, seonkim\}@usc.edu, tripwithsoo@gmail.com,
abdullah.alfarrarjeh@gju.edu.jo
}
\thanks{\textsuperscript{*}Corresponding author: Seon Ho Kim.}
}

\maketitle
\begin{abstract}
Large collections of street-view imagery provide rich visual information about urban environments, but fine-grained regional geolocalization remains challenging because nearby areas often share coarse geographic cues. We study regional geolocalization within a metropolitan
area and ask whether pretrained CLIP features are sufficient for
regional discrimination and which visual information supports
performance after adaptation. Using 9,085 street-view images from eight Greater Los Angeles regions, we compare zero-shot CLIP, frozen-encoder readouts, and encoder adaptation. Frozen readouts remain near the 39.03\% zero-shot baseline, whereas encoder adaptation achieves 75.94--82.10\% accuracy. We examine these gains through semantic cue removal, edge and blur transformations, and patch scrambling. Adapted models achieve higher accuracy on appearance-reduced images but do not retain a larger fraction of their original accuracy. They also change predictions more frequently under patch scrambling, while vegetation and sky remain influential. A zero-shot Caltech101 control shows that scrambling also disrupts generic object recognition. Together, these findings support increased sensitivity to scene configuration alongside continued sensitivity to appearance cues, without establishing that coarse structure alone is sufficient for prediction.
\end{abstract}

\begin{IEEEkeywords}
regional geolocalization, CLIP adaptation, scene configuration,
visual interventions, vision--language models
\end{IEEEkeywords}

\section{Introduction}

Visual geolocalization aims to infer where an image was captured from its visual content, supporting applications such as navigation, image organization, and geographic information retrieval. Existing approaches use image retrieval, coordinate estimation, or geographic classification \cite{hays2008im2gps,vo2017im2gps,cepeda2023geoclip}. We select CLIP, a contrastive vision--language model \cite{radford2021clip}, because it enables controlled comparisons of zero-shot prediction, frozen-encoder baselines, and encoder adaptation within a shared architecture without language generation. CLIP-based methods, including StreetCLIP, GeoCLIP, PIGEON, and AddressCLIP, have demonstrated promising results in geolocalization at different geographic scales\cite{haas2023streetclip,cepeda2023geoclip,haas2024pigeon,xu2024addressclip}.

However, distinguishing nearby places within one metropolitan area remains challenging. Broad geographic cues, including language, vegetation, road systems, and architectural styles, may be shared across neighboring areas. Regional discrimination must therefore draw on subtle differences in streetscape appearance and the arrangement of scene elements. This setting raises questions about both the regional information accessible through pretrained representations and the visual evidence associated with improvements after adaptation.

\begin{figure}[!t]
\centering
\includegraphics[width=\columnwidth]{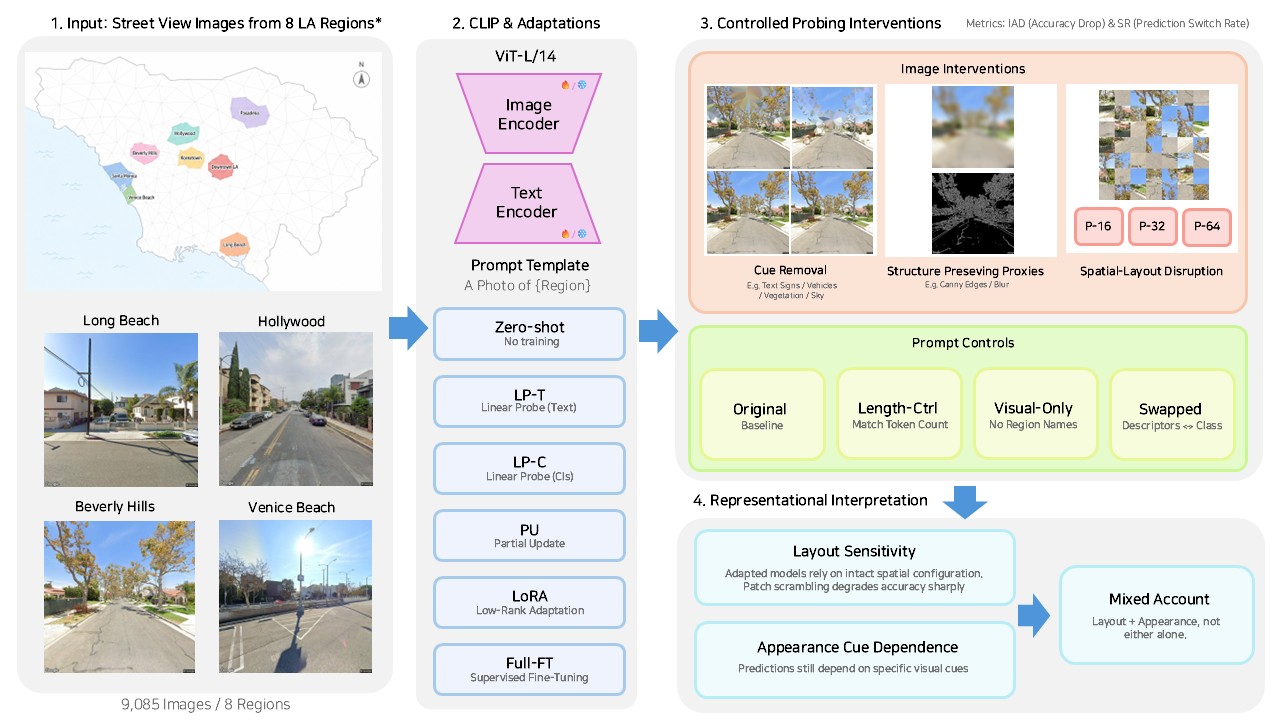}
\caption{Overview: regional geolocalization with CLIP
adaptation, visual interventions, and zero-shot prompt controls.}
\label{fig:overview}
\end{figure}

We study \emph{regional geolocalization}, defined as classification among selected areas within a single metropolitan region. Our dataset contains 9,085 street-view images from eight Greater Los Angeles municipalities, neighborhoods, and districts, which we collectively call regions. It provides a focused setting for examining discrimination among nearby areas with overlapping visual characteristics.

Our first question is: \emph{How effectively can frozen CLIP readouts distinguish these regions, and how much does adapting the image encoder improve performance?} We compare zero-shot CLIP, frozen-encoder text-score (LP-T) and linear-classifier (LP-C) readouts, Partial Update, Low-Rank Adaptation (LoRA) \cite{hu2022lora}, and full fine-tuning (Full-FT). This comparison separates learning an output mapping from modifying the visual representation. Frozen readouts remain near the 39.03\% zero-shot baseline, whereas encoder adaptation achieves 75.94--82.10\% accuracy. These results show that the evaluated encoder-adaptation methods recover regional discrimination beyond that achieved by the frozen readouts.

These gains motivate our second question: \emph{How does sensitivity to visual information change after adaptation?} Accuracy alone does not explain how a model distinguishes nearby regions. Scenes can contain similar buildings, roads, vehicles, and vegetation, yet differ in their appearance and arrangement. We therefore examine whether adaptation changes sensitivity to selected scene cues, detailed appearance, and scene configuration, defined here as the arrangement of visual elements within an image.

To examine this question, we use three families of input interventions: \emph{semantic cue removal}, \emph{appearance reduction}, and \emph{scene-layout disruption}. Together, they probe prediction changes when selected visual cues, detailed appearance, or the arrangement of local content is altered. We interpret these interventions jointly because each changes multiple image properties. Figure~\ref{fig:overview} summarizes the study design.

After adaptation, predictions change more frequently when scene layout is disrupted, while environmental cues such as vegetation and sky remain influential. Although accuracy on appearance-reduced images is higher, the fraction of original performance retained does not improve. Together, these observations associate encoder adaptation with greater dependence on intact scene configuration, without evidence that coarse structure alone becomes sufficient for prediction.

We make three main contributions:
\begin{itemize}
    \item We establish a substantial gap between frozen readouts
    and encoder adaptation for regional geolocalization in
    Greater Los Angeles.
    \item We provide a complementary intervention-based analysis of the visual information associated with adaptation gains.
    \item We distinguish sensitivity to scene configuration from
    the sufficiency of coarse structure and identify continued
    sensitivity to environmental appearance cues.
\end{itemize}

The evaluation includes same-coordinate and nearby views across training and test sets. Our findings therefore concern viewpoint variation near known locations.

\section{Related Work}

\subsection{Visual Geolocalization}

Visual geolocalization uses classification, retrieval, coordinate prediction, or combinations thereof. PlaNet learns a classifier over geographic cells \cite{weyand2016planet}; Deep Img2GPS combines classification features with retrieval and density estimation \cite{vo2017im2gps}; and data-centric work examines image scene localization \cite{alfarrarjeh2018datacentric}. OSV-5M emphasizes global coverage and strict geographic train--test separation \cite{astruc2024osv5m}. Our independently collected LA dataset does not use OSV-5M images; OSV-5M instead provides a methodological precedent for geographic separation.

\subsection{CLIP Adaptation for Geolocalization}

CLIP provides transferable image--text representations \cite{radford2021clip}. StreetCLIP studies open-domain zero-shot geolocalization \cite{haas2023streetclip}; GeoCLIP aligns images with continuous GPS representations \cite{cepeda2023geoclip}; PIGEON combines semantic geocells, contrastive pretraining, and retrieval refinement \cite{haas2024pigeon}; and AddressCLIP studies city-wide address prediction \cite{xu2024addressclip}. These works establish CLIP-based geographic prediction. Our focus is not a new city-scale application, but the visual information associated with adaptation gains under controlled interventions; predictive performance alone does not identify that information.

\subsection{Probing Object Relations and Spatial Arrangement}

Downstream accuracy does not by itself imply relational understanding. The Attribution, Relation, and Order (ARO) benchmark probes attributes, relations, and word order \cite{yuksekgonul2023aro}, while SpatialCLIP targets broader 3-D-inspired spatial understanding \cite{wang2025spatialclip}. Our interventions test only coarse scene-configuration sensitivity and selected appearance-cue dependence; no individual corruption establishes general spatial reasoning. This motivates a controlled setting in which adaptation and prediction sensitivity can be examined together.

\section{Task and Dataset}

\subsection{Los Angeles-Specific Dataset and Operational Definition}

\begin{figure}[!t]
\centering
\includegraphics[width=\columnwidth]{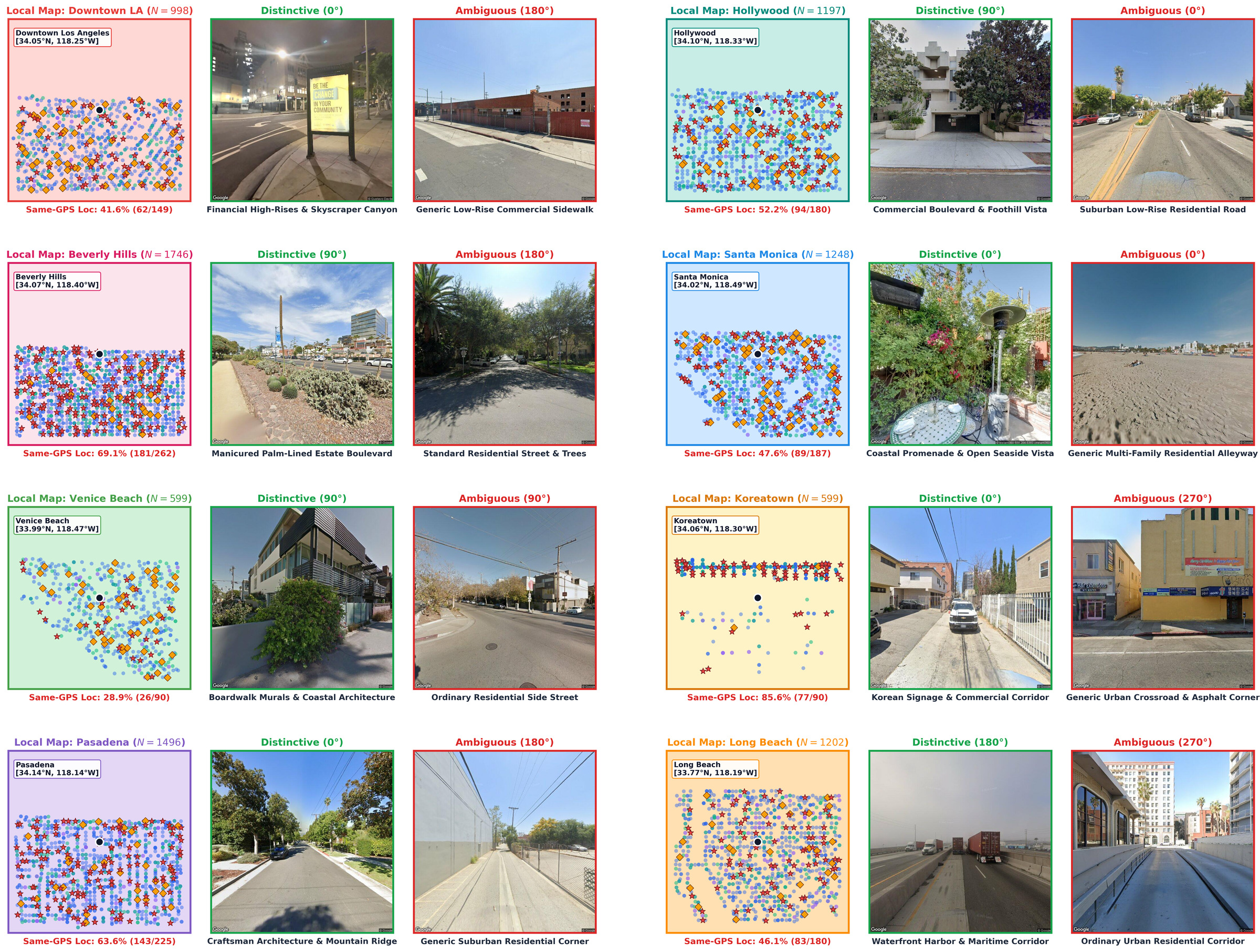}
\caption{Sampled locations and distinctive and ambiguous views
from eight Greater Los Angeles regions. $N$: regional sample
size; Same-GPS Loc: test images sharing a requested coordinate
with training.}
\label{fig:dataset-overview}
\end{figure}

We construct a Los Angeles street-view dataset through the Google Maps API; it is not derived from OSV-5M \cite{astruc2024osv5m}. Let $\mathcal{D}=\{(x_i,y_i,g_i)\}_{i=1}^{N_{\mathcal{D}}}$ denote a dataset of images \(x_i\), region labels \(y_i\) from \(K=8\) classes, and geographic coordinates \(g_i\), with $N_{\mathcal{D}}=9{,}085$. The classes are Downtown Los Angeles, Pasadena, Beverly Hills, Santa Monica, Hollywood, Long Beach, Venice Beach, and Koreatown (Figure \ref{fig:dataset-overview}). Four are incorporated cities and the others are recognized neighborhoods or districts; we collectively call them regions and formulate the task as sub-regional classification within Greater Los Angeles.

Let $\mathcal{E}$ denote the evaluation set, $N_{\mathcal{E}}=|\mathcal{E}|$, and $\hat{y}_{i,m}$ the prediction of model $m$ for image $x_i$. Top-1 accuracy is:
\begin{equation}
\operatorname{Acc}(m)=\frac{1}{N_{\mathcal{E}}}\sum_{i\in\mathcal{E}}\mathbb{I}\!\left[\hat{y}_{i,m}=y_i\right].
\end{equation}
Let $c_k$ denote the geographic centroid of region $k$ and $d(g_i,c_k)$ the distance in kilometers between image coordinate $g_i$ and that centroid. We define centroid error as the mean distance to the centroid of the predicted region:
\begin{equation}
\operatorname{CE}(m)=\frac{1}{N_{\mathcal{E}}}\sum_{i\in\mathcal{E}}d(g_i,c_{\hat{y}_{i,m}}).
\end{equation}

\subsection{Dataset Reconstruction and Density-Based Sampling}

Of the 28,550 raw images, 30 were outside the target regions, leaving 28,520 eligible images. Let $n_k$ and $A_k$ denote the number of eligible images and area of region $k$, respectively. To reduce regional density imbalance, we sample at the largest common density, $\rho^{\star}=\min_k(n_k/A_k)=194.49$ images/km$^2$, determined by Koreatown. Sampling approximately $\rho^{\star}A_k$ images per region
yields the final dataset of 9,085 images. This equalizes spatial density rather than class counts;
larger regions contribute more images.

\subsection{Dataset Splits, Integrity, and Spatial Overlap}
\label{sec:dataset-splits}

We use a region-stratified 70/15/15 split: 6,359 training, 1,363 validation, and 1,363 test images, with seed 42. We retain split metadata and raw images for reproducibility.

\subsubsection{Data Collection and Processing}

Available metadata primarily indicates that most images were collected in early 2026. At each coordinate, the API returned separate perspective images at $0^{\circ}$, $90^{\circ}$, $180^{\circ}$, and $270^{\circ}$ rather than stitched panoramas. Polygons define region membership, and coordinate/heading metadata supports the overlap audit. We applied no geographic or embedding-based deduplication;
density normalization cannot replace coordinate-grouped splitting or duplicate filtering.

\subsubsection{Spatial Overlap and Viewpoint Generalization}

The image-level split is geographically overlapping: among all 1,363 test records, 755 (55.39\%) share a requested coordinate with training and 1,086 (79.68\%) lie within 50~m of a training coordinate; the remaining 277 (20.32\%) are separated by more than 50~m. Orthogonal headings can show different fields of view---for example, a residential street versus a commercial cross-street---and reduce direct pixel correspondence, yet retain local buildings, intersection geometry, vegetation, road surfaces, and capture conditions. The protocol therefore measures viewpoint variation near known locations, not geographically disjoint generalization. Correct classification from an unseen heading may reflect viewpoint tolerance but cannot exclude micro-location memorization. Coordinate-grouped or spatially buffered splits are required for stricter evaluation; OSV-5M provides a methodological contrast \cite{astruc2024osv5m}.
\section{Adaptation Methods}

\subsection{Common Training Configuration}

All methods use OpenCLIP's pretrained \texttt{ViT-L-14}
image encoder~\cite{dosovitskiy2021image} with its default
$224\times224$ resize, center crop, and normalization,
without stochastic augmentation. Supervised methods train
for 10 epochs with seed 42, batch size 8, AdamW, weight
decay 0.01, cosine annealing ($T_{\max}=10$), and automatic
mixed precision with dynamic loss scaling. Batches are
shuffled without weighted sampling, and cross-entropy loss
is unweighted. We evaluate the checkpoint with the highest
validation accuracy. Learning rates are $10^{-4}$ for
LP-T, LP-C, and LoRA, and $10^{-5}$ for PU and Full-FT.

\subsection{Adaptation Strategies}

Zero-shot CLIP predicts from image--text similarity using
the template ``a Google Street View photo in \{Region\},
Los Angeles.'' LP-T learns only the scale and bias of
text-based class scores, whereas LP-C trains a linear
classifier over frozen image features. Partial Update (PU)
updates the final visual transformer block, \texttt{ln\_post},
and the classifier. LoRA trains rank-8 adapters
($\alpha=16$, dropout 0.1) in attention and MLP projections
while freezing the pretrained encoder
weights~\cite{hu2022lora}. Full-FT
updates the entire image encoder and classifier.

\section{Intervention-Based Probing Protocol}

We jointly interpret cue removal, appearance reduction, and layout disruption because each changes multiple image properties, reporting transformed accuracy and change relative to original-image performance.

\subsection{Cue Selection and Removal}
\label{sec:cue-removal}

We select text and vehicles as semantic or activity-related
cues, and vegetation and sky as environmental cues.
One-way ANOVA shows regional differences in mask coverage
for all four cues (all $p<0.01$), with the largest
$F$ statistics for vegetation ($F=35.04$) and sky
($F=10.93$). This variation motivates testing their
predictive relevance through cue removal. We exclude roads and buildings because extensive inpainting
could alter scene geometry.

Text regions are detected using EasyOCR's CRAFT
detector~\cite{baek2019craft}. Vehicles are detected using
MS-COCO-pretrained YOLOv8n~\cite{jocher2023yolov8},
while vegetation and sky masks use ADE20K-pretrained
SegFormer-B0~\cite{xie2021segformer}.
Telea inpainting~\cite{telea2004inpainting} fills removed
regions. For each cue mask, we generate a random mask
with the same pixel count to control for removed area.

For transformation $T$, let $\hat{y}_{i,m}^{(T)}$ denote
the prediction on $T(x_i)$ and $\operatorname{Acc}(m;T)$
its accuracy. For cue $q$ and removal $T_q$,
intervention-induced accuracy drop (IAD) is:
\begin{equation}
\operatorname{IAD}(m,q)=\operatorname{Acc}(m)-\operatorname{Acc}(m;T_q),
\end{equation}
and prediction switch rate:
\begin{equation}
\operatorname{SR}(m,q)=\frac{1}{N_{\mathcal{E}}}\sum_{i\in\mathcal{E}}\mathbb{I}\!\left[\hat{y}_{i,m}\neq\hat{y}_{i,m}^{(T_q)}\right].
\end{equation}
SR is the fraction of labels changed by removal and measures instability rather than correctness \cite{hendrycks2019benchmarking}. The cue-removal experiment therefore characterizes sensitivity to the implemented removals rather than causal cue importance.

\subsection{Edge and Blur Transformations}

Canny edges \cite{canny1986edge} and Gaussian macro blur reduce fine appearance while approximately preserving coarse organization. They do not isolate structure: edge images retain text outlines, vehicle shapes, windows, and vegetation boundaries, while blur retains color and coarse texture. Performance therefore cannot be attributed exclusively to spatial structure. Alongside transformed accuracy, we report retention, which normalizes by each model's original accuracy:
\begin{equation}
\operatorname{Retention}(m,T)=\frac{\operatorname{Acc}(m;T)}{\operatorname{Acc}(m)}.
\end{equation}

\subsection{Spatial-Layout Disruption and Prompt Controls}

Patch scrambling randomly permutes a regular image grid, preserving block content while disrupting global arrangement; Patch-$n$ uses $n\times n$-pixel blocks. Larger blocks retain more coherent local content. Scrambling also introduces unnatural boundaries, object fragmentation, distribution shift, and interactions with positional embeddings. Accordingly, we interpret prediction changes as layout sensitivity rather than semantic spatial reasoning or uniquely geographic structure use.

For zero-shot CLIP, we compare the original prompts with three variants: length-controlled prompts with neutral filler, visual-only descriptions of architecture, vegetation, and streetscapes without region names, and descriptor-swapped prompts that cyclically reassign descriptions across regions. These controls assess dependence on prompt wording and descriptor--label associations.

\begin{figure*}[t]
\centering
\includegraphics[width=\textwidth]{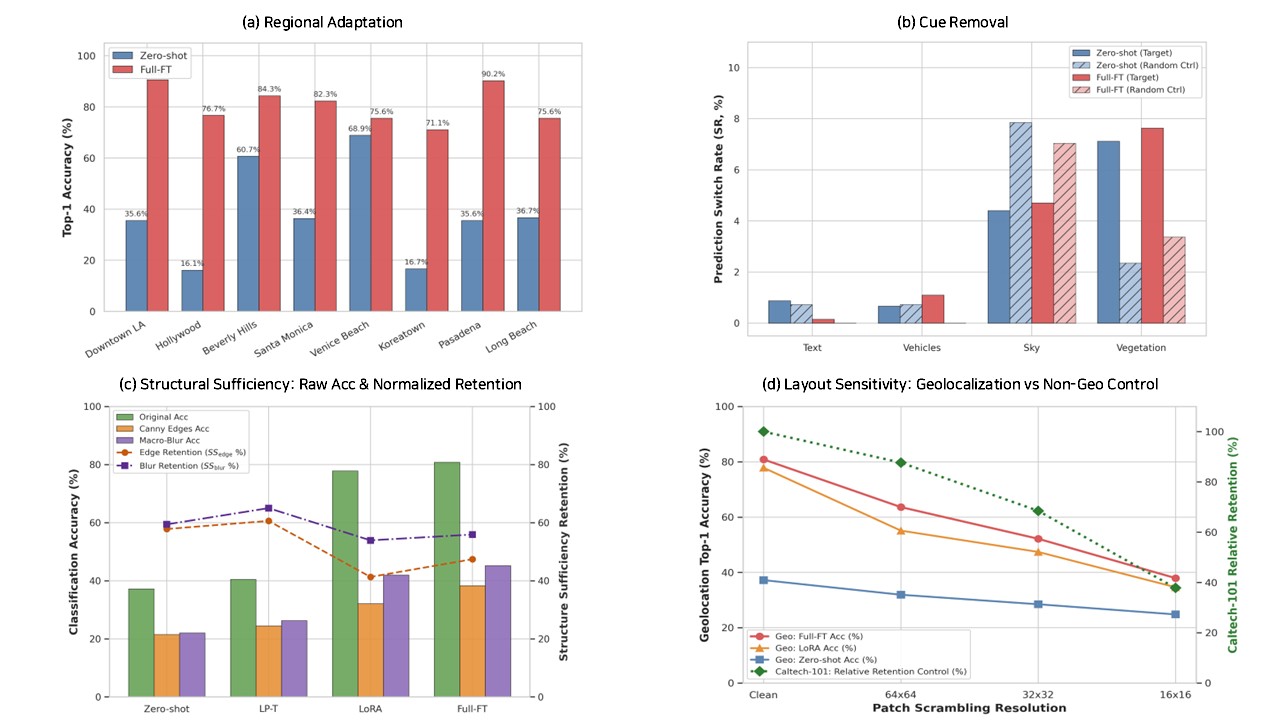}
\caption{Intervention summary: (a) regional gains, (b) cue removal and matched random masks, (c) accuracy and retention after appearance reduction, and (d) layout disruption in geolocalization and Caltech101. Panels (c) and (d) contrast structural sufficiency and sensitivity. Acc: accuracy; Ctrl: control; Geo: geolocalization; SS: structural-sufficiency retention.}
\label{fig:probe-summary}
\end{figure*}

\section{Results and Discussion}

\subsection{Overall Adaptation Performance}

\begin{table}[t]
\caption{Original-image performance, zero-shot prompt controls,
and prediction switch rates under patch scrambling.}
\label{tab:adaptation}
\label{tab:patch}
\centering
\small
\setlength{\tabcolsep}{3pt}
\begin{tabular}{@{}lccccc@{}}
\toprule
\multirow{2}{*}{Setting}
& \multicolumn{2}{c}{Original images}
& \multicolumn{3}{c}{Scrambling SR (\%)} \\
\cmidrule(lr){2-3}\cmidrule(lr){4-6}
& Acc. (\%) & CE (km) & 16 & 32 & 64 \\
\midrule
Zero-shot (original)
& 39.03 & 12.30 & 12.40 & 8.73 & 5.28 \\
LP-T
& 41.45 & 12.25 & 14.60 & 9.24 & 6.46 \\
LP-C
& 38.52 & 11.22 & 10.79 & 7.92 & 4.92 \\
PU
& 75.94 & 5.08 & 45.56 & 41.31 & 36.10 \\
LoRA
& 78.36 & 4.21 & 43.21 & 30.45 & 22.74 \\
Full-FT
& \textbf{82.10} & \textbf{3.86} & 42.92 & 28.69 & 17.24 \\
\midrule
\multicolumn{6}{@{}l}{\textit{Zero-shot prompt controls}} \\
Length-controlled
& 38.30 & -- & -- & -- & -- \\
Visual-only
& 36.24 & -- & -- & -- & -- \\
Descriptor-swapped
& 11.15 & -- & -- & -- & -- \\
\bottomrule
\end{tabular}
\end{table}

Encoder adaptation achieves 75.94--82.10\% accuracy, whereas frozen readouts remain near the 39.03\% zero-shot baseline (Table~\ref{tab:adaptation}). Full-FT also reduces centroid error from 12.30 to 3.86~km. Figure~\ref{fig:probe-summary}(a) shows improvements across all eight regions, ranging from 6.67 percentage points in Venice Beach to 60.56 points in Hollywood. These differences indicate that adaptation benefits vary substantially across regions. Under the evaluated readouts and overlapping split, encoder updating provides substantially better regional discrimination. Low frozen-probe accuracy does not imply that CLIP lacks regional information.

For zero-shot CLIP, length-controlled and visual-only prompts yield accuracies close to the original baseline, whereas descriptor swapping reduces accuracy to 11.15\% (Table~\ref{tab:adaptation}). This control indicates sensitivity to descriptor--label associations, but does not establish visual grounding of individual descriptors.

\subsection{Cue-Removal Sensitivity}

Vegetation and sky removal change more predictions than text or vehicle removal (Figure~\ref{fig:probe-summary}(b)). For LoRA, switch rates are 7.92\%, 5.65\%, 1.83\%, and 1.91\% for vegetation, sky, text, and vehicles, respectively. The matched random-mask controls provide a reference for the effect of removing image area, while differences in mask coverage and segmentation quality limit causal ranking across cue categories.

\subsection{Appearance Reduction and Layout Sensitivity}

LoRA and Full-FT achieve higher accuracy on edge-transformed and blurred images, but do not retain a larger fraction of their original performance (Table~\ref{tab:structure}; Figure~\ref{fig:probe-summary}(c)). For example, Full-FT raises edge accuracy from 21.5\% to 38.3\%, while retention decreases from 0.55 to 0.47. This distinction indicates that adapted models make more correct predictions from the information remaining after transformation, without preserving a greater proportion of their original accuracy. Thus, higher transformed accuracy alone does not establish that coarse structure is sufficient for prediction.

\begin{figure*}[!t]
\centering
\includegraphics[width=\textwidth]{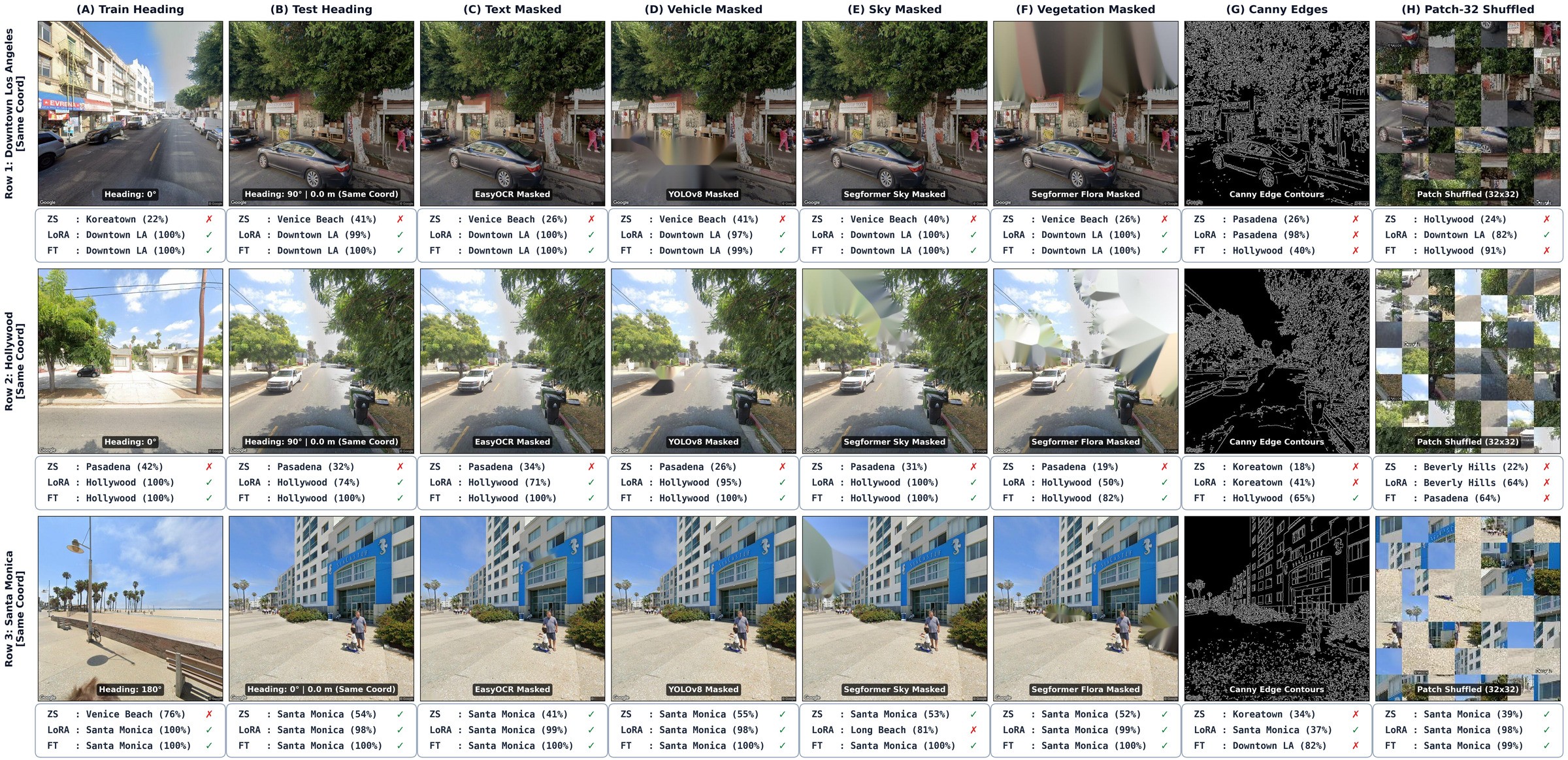}
\caption{Qualitative cases from Downtown Los Angeles, Hollywood, and Santa Monica. Columns show training and test viewing directions, four cue masks, Canny edge maps, and Patch-32 scrambling, together with predictions from zero-shot CLIP, LoRA, and Full-FT. 
}
\label{fig:qualitative}
\end{figure*}

Under Patch-16, adapted models switch 42.92--45.56\% of predictions, compared with 10.79--14.60\% for frozen methods (Table~\ref{tab:patch}). The gap persists at larger patch sizes, although switching decreases as larger blocks preserve more coherent local content. This within-task contrast is consistent with greater sensitivity to intact scene configuration after adaptation. Figure~\ref{fig:qualitative} illustrates intervention and viewpoint cases.

\begin{table}[t]
\caption{Accuracy and relative retention on appearance-reduced structure proxies.}
\label{tab:structure}
\centering
\small
\begin{tabular}{llcc}
\toprule
Input & Model & Accuracy (\%) & Retention \\
\midrule
\multirow{3}{*}{Edges}
 & Zero-shot & 21.5 & 0.55 \\
 & LoRA & 32.1 & 0.41 \\
 & Full-FT & 38.3 & 0.47 \\
\midrule
\multirow{3}{*}{Macro blur}
 & Zero-shot & 22.1 & 0.57 \\
 & LoRA & 42.0 & 0.54 \\
 & Full-FT & 45.2 & 0.55 \\
\bottomrule
\end{tabular}
\end{table}

\subsection{Non-Geographic Control on Caltech101}

We apply the same scrambling implementation to 2,000 Caltech101 images using seed-42 and resized to $224\times224$ \cite{feifei2004caltech101}. Sample indices are recorded for reproducibility. The zero-shot pipeline uses 102 labels, including the loader's background label, and no Caltech101 training data. For the fixed model $m$, relative drop is:
\begin{equation}
\operatorname{Drop}(m,T)=\frac{\operatorname{Acc}(m)-\operatorname{Acc}(m;T)}{\operatorname{Acc}(m)}\times100.
\end{equation}

\begin{table}[t]
\caption{Caltech101 zero-shot accuracy under the same patch-scrambling protocol.}
\label{tab:caltech-control}
\centering
\small
\begin{tabular}{lcc}
\toprule
Condition & Accuracy (\%) & Relative drop (\%) \\
\midrule
Clean & 89.65 & 0.00 \\
Patch-16 & 33.90 & 62.19 \\
Patch-32 & 61.40 & 31.51 \\
Patch-64 & 78.60 & 12.33 \\
\bottomrule
\end{tabular}
\end{table}

Caltech101 accuracy drops by 62.19\%, 31.51\%, and 12.33\% under Patch-16, -32, and -64 (Table~\ref{tab:caltech-control}). Scrambling therefore disrupts generic recognition and cannot by itself establish geolocalization-specific sensitivity. These accuracy drops are not directly comparable to geolocalization SR, which measures any label change. The within-task frozen--adapted comparison supports increased
layout sensitivity.

\section{Limitations}

The study covers one metropolitan area and eight comprising municipality/neighborhood region types, so it does not establish behavior in other cities, scales, or image sources. Its overlapping split evaluates viewpoint variation near known locations, not geographic generalization. Edge and blur are imperfect structure proxies, scrambling introduces artifacts, and cue removal depends on mask area and segmentation quality. Model-performance comparisons use a single seed and are not accompanied by significance tests or confidence intervals. The study also lacks human comparison. Street View images cannot be redistributed \cite{google2026maps}. Future work should use coordinate-grouped or buffered splits, matched and audited masks, alternative fills, multiple scrambling permutations, aligned accuracy-drop and SR metrics, multiple seeds, stronger frozen-feature probes, and cross-source, temporal, seasonal, and geographically disjoint evaluation.

\section{Conclusion}

Within these limitations, our experiments provide consistent evidence about the effects of CLIP adaptation on regional geolocalization. Frozen readouts remain near the 39.03\% zero-shot baseline, whereas encoder adaptation reaches 75.94--82.10\%; Full-FT also reduces centroid error from 12.30~km to 3.86~km. Under the evaluated setting, strong nearby-region discrimination therefore emerges primarily when the pretrained visual representation is allowed to adapt.

Adapted models also achieve higher edge and blur accuracy and switch more often under layout disruption, but do not retain a larger fraction of original accuracy after appearance reduction. Vegetation and sky remain influential, and Caltech101 shows that scrambling sensitivity is not uniquely geolocalization-specific. Together, the results support greater dependence on intact scene configuration without establishing structural sufficiency; predictions remain consistent with a combination of configuration and appearance cues. Within this scope, regional geolocalization provides a controlled setting for examining how adaptation changes visual evidence, but the findings concern viewpoint variation near known locations and require confirmation with geographically separated evaluation.

\section*{Acknowledgment}
This research has been supported in part by the USC Integrated Media Systems Center. OpenAI GPT-5.6~\cite{openai2026gpt56} and Google Gemini 3.1 Pro~\cite{googledeepmind2026gemini31} assisted with English translation, language and style editing, code, and figures. The human authors reviewed and approved all AI-assisted content. 

\bibliographystyle{IEEEtran}
\bibliography{references}

@inproceedings{radford2021clip,
  author    = {Radford, Alec and Kim, Jong Wook and Hallacy, Chris and Ramesh, Aditya and Goh, Gabriel and Agarwal, Sandhini and Sastry, Girish and Askell, Amanda and Mishkin, Pamela and Clark, Jack and Krueger, Gretchen and Sutskever, Ilya},
  title     = {Learning Transferable Visual Models From Natural Language Supervision},
  booktitle = {Proc. {ICML}},
  pages     = {8748--8763},
  year      = {2021}
}

@inproceedings{dosovitskiy2021image,
  author    = {Dosovitskiy, Alexey and Beyer, Lucas and Kolesnikov, Alexander and Weissenborn, Dirk and Zhai, Xiaohua and Unterthiner, Thomas and Dehghani, Mostafa and Minderer, Matthias and Heigold, Georg and Gelly, Sylvain and Uszkoreit, Jakob and Houlsby, Neil},
  title     = {An Image Is Worth 16x16 Words: Transformers for Image Recognition at Scale},
  booktitle = {Proc. {ICLR}},
  year      = {2021}
}

@inproceedings{weyand2016planet,
  author    = {Weyand, Tobias and Kostrikov, Ilya and Philbin, James},
  title     = {{PlaNet}: Photo Geolocation with Convolutional Neural Networks},
  booktitle = {Proc. {ECCV}},
  pages     = {37--55},
  year      = {2016}
}

@inproceedings{hays2008im2gps,
  author    = {Hays, James and Efros, Alexei A.},
  title     = {{IM2GPS}: Estimating Geographic Information from a Single Image},
  booktitle = {Proc. {CVPR}},
  pages     = {1--8},
  year      = {2008}
}

@inproceedings{vo2017im2gps,
  author    = {Vo, Nam and Jacobs, Nathan and Hays, James},
  title     = {Revisiting {IM2GPS} in the Deep Learning Era},
  booktitle = {Proc. {ICCV}},
  pages     = {2621--2630},
  year      = {2017}
}

@inproceedings{alfarrarjeh2018datacentric,
  author    = {Alfarrarjeh, Abdullah and Kim, Seon Ho and Rajan, Shivnesh and Deshmukh, Akshay and Shahabi, Cyrus},
  title     = {A Data-Centric Approach for Image Scene Localization},
  booktitle = {Proc. {IEEE} Big Data},
  pages     = {594--603},
  year      = {2018},
  doi       = {10.1109/BigData.2018.8621912}
}

@article{haas2023streetclip,
  author  = {Haas, Lukas and Alberti, Silas and Skreta, Marta},
  title   = {Learning Generalized Zero-Shot Learners for Open-Domain Image Geolocalization},
  journal = {arXiv preprint arXiv:2302.00275},
  year    = {2023}
}

@inproceedings{cepeda2023geoclip,
  author    = {Vivanco Cepeda, Vicente and Nayak, Gaurav Kumar and Shah, Mubarak},
  title     = {{GeoCLIP}: {CLIP}-Inspired Alignment between Locations and Images for Effective Worldwide Geo-localization},
  booktitle = {Proc. {NeurIPS}},
  volume    = {36},
  pages     = {8690--8701},
  year      = {2023}
}

@inproceedings{haas2024pigeon,
  author    = {Haas, Lukas and Skreta, Michal and Alberti, Silas and Finn, Chelsea},
  title     = {{PIGEON}: Predicting Image Geolocations},
  booktitle = {Proc. {CVPR}},
  pages     = {12893--12902},
  year      = {2024}
}

@inproceedings{xu2024addressclip,
  author    = {Xu, Shixiong and Zhang, Chenghao and Fan, Lubin and Meng, Gaofeng and Xiang, Shiming and Ye, Jieping},
  title     = {{AddressCLIP}: Empowering Vision-Language Models for City-wide Image Address Localization},
  booktitle = {Proc. {ECCV}},
  pages     = {76--92},
  year      = {2024}
}

@inproceedings{astruc2024osv5m,
  author    = {Astruc, Guillaume and Dufour, Nicolas and Siglidis, Ioannis and Aronssohn, Constantin and Bouia, Nacim and Fu, Stephanie and Loiseau, Romain and Nguyen, Van Nguyen and Raude, Charles and Vincent, Elliot and Xu, Lintao and Zhou, Hongyu and Landrieu, Loic},
  title     = {{OpenStreetView-5M}: The Many Roads to Global Visual Geolocation},
  booktitle = {Proc. {CVPR}},
  pages     = {21967--21977},
  year      = {2024}
}

@inproceedings{yuksekgonul2023aro,
  author    = {Yuksekgonul, Mert and Bianchi, Federico and Kalluri, Pratyusha and Jurafsky, Dan and Zou, James},
  title     = {When and Why Vision-Language Models Behave Like Bags-of-Words, and What to Do About It?},
  booktitle = {Proc. {ICLR}},
  year      = {2023}
}

@inproceedings{wang2025spatialclip,
  author    = {Wang, Zehan and Zhou, Sashuai and He, Shaoxuan and Huang, Haifeng and Yang, Lihe and Zhang, Ziang and Cheng, Xize and Ji, Shengpeng and Jin, Tao and Zhao, Hengshuang and Zhao, Zhou},
  title     = {{SpatialCLIP}: Learning 3D-Aware Image Representations from Spatially Discriminative Language},
  booktitle = {Proc. {CVPR}},
  pages     = {29656--29666},
  year      = {2025}
}

@inproceedings{hu2022lora,
  author    = {Hu, Edward J. and Shen, Yelong and Wallis, Phillip and Allen-Zhu, Zeyuan and Li, Yuanzhi and Wang, Shean and Wang, Lu and Chen, Weizhu},
  title     = {{LoRA}: Low-Rank Adaptation of Large Language Models},
  booktitle = {Proc. {ICLR}},
  year      = {2022}
}

@inproceedings{feifei2004caltech101,
  author    = {Li, Fei-Fei and Fergus, Rob and Perona, Pietro},
  title     = {{Learning Generative Visual Models from Few Training Examples: An Incremental Bayesian Approach Tested on 101 Object Categories}},
  booktitle = {Proc. {CVPR} Workshop Generative-Model Based Vis.},
  year      = {2004}
}

@inproceedings{hendrycks2019benchmarking,
  author    = {Hendrycks, Dan and Dietterich, Thomas},
  title     = {Benchmarking Neural Network Robustness to Common Corruptions and Perturbations},
  booktitle = {Proc. {ICLR}},
  year      = {2019}
}

@inproceedings{baek2019craft,
  author    = {Baek, Youngmin and Lee, Bado and Han, Dongyoon and Yun, Sangdoo and Lee, Hwalsuk},
  title     = {Character Region Awareness for Text Detection},
  booktitle = {Proc. {CVPR}},
  pages     = {9365--9374},
  year      = {2019}
}

@misc{jocher2023yolov8,
  author       = {Jocher, Glenn and Chaurasia, Ayush and Qiu, Jing},
  title        = {{Ultralytics YOLOv8}},
  year         = {2023},
  howpublished = {Software, version 8.0.0},
  note         = {Available: \url{https://github.com/ultralytics/ultralytics}}
}

@inproceedings{xie2021segformer,
  author    = {Xie, Enze and Wang, Wenhai and Yu, Zhiding and Anandkumar, Anima and Alvarez, Jose M. and Luo, Ping},
  title     = {{SegFormer}: Simple and Efficient Design for Semantic Segmentation with Transformers},
  booktitle = {Proc. {NeurIPS}},
  volume    = {34},
  pages     = {12077--12090},
  year      = {2021}
}

@article{telea2004inpainting,
  author  = {Telea, Alexandru},
  title   = {An Image Inpainting Technique Based on the Fast Marching Method},
  journal = {J. Graphics Tools},
  volume  = {9},
  number  = {1},
  pages   = {23--34},
  year    = {2004}
}

@article{canny1986edge,
  author  = {Canny, John},
  title   = {A Computational Approach to Edge Detection},
  journal = {IEEE Trans. Pattern Anal. Mach. Intell.},
  volume  = {8},
  number  = {6},
  pages   = {679--698},
  year    = {1986}
}

@misc{google2026maps,
  author       = {{Google LLC}},
  title        = {{Google Maps Platform Terms of Service}},
  year         = {2026},
  howpublished = {\url{https://cloud.google.com/maps-platform/terms}},
  note         = {Accessed: Aug. 14, 2026}
}

@misc{openai2026gpt56,
  author       = {{OpenAI}},
  title        = {{GPT}-5.6 System Card},
  year         = {2026},
  howpublished = {OpenAI Deployment Safety Hub, \url{https://deploymentsafety.openai.com/gpt-5-6}},
  note         = {Published: July 9, 2026; accessed: Aug. 21, 2026}
}

@misc{googledeepmind2026gemini31,
  author       = {{Google DeepMind}},
  title        = {{Gemini 3.1 Pro Model Card}},
  year         = {2026},
  howpublished = {\url{https://deepmind.google/models/model-cards/gemini-3-1-pro/}},
  note         = {Published: Feb. 19, 2026; accessed: Aug. 21, 2026}
}

\end{document}